%% file: main.tex
\documentclass[letterpaper, 10 pt, conference]{ieeeconf}
\IEEEoverridecommandlockouts

\usepackage{url}            
\usepackage{booktabs}       
\usepackage{nicefrac}       
\usepackage{microtype}      
\usepackage{amsmath}
\usepackage{amsfonts}
\usepackage{amssymb}
\usepackage{graphicx} 
\usepackage{algorithm}
\usepackage{algpseudocode}
\usepackage{xcolor}

\usepackage{tabularx}  
\usepackage{booktabs}  
\usepackage{graphicx}  
\usepackage{fix-cm}
\usepackage{hyperref}

\begin{document}

\title{Bayesian Retraction Optimization for \\ Tissue Attachment Mapping in Surgical Dissection}

\author{
Shing-Hei~Ho$^{1,2}$,
Bao~Thach$^{1}$,
Toan~Vo$^{1}$,
James~M.~Ferguson$^{3}$, and 
Alan~Kuntz$^{3}$%
\thanks{
Research reported in this publication was supported by the Advanced Research Projects Agency for Health (ARPA-H) under the ALISS project, Award Number D24AC00415-00. 
The ARPA-H award of up to \$11,935,038 provided 100\% of the financial support for this work. 
The opinions and findings in this paper are solely the responsibility of the authors and do not necessarily represent the official views of ARPA-H.
}%
\thanks{
$^{1}$Work was performed by authors at The Robotics Center and the Kahlert School of Computing at the University of Utah, Salt Lake City, UT, USA.
}%
\thanks{
$^{2}$Shing-Hei Ho is with the School of Computational Science and Engineering, Georgia Institute of Technology, GA, USA.
}%
\thanks{
$^{3}$James Ferguson and Alan Kuntz are with the Department of Electrical and Computer Engineering and Department of Computer Science, Vanderbilt University, TN, USA.
{\tt\footnotesize james.m.ferguson@vanderbilt.edu}}}

\maketitle

\begin{abstract}

With growing surgeon shortages, automating surgical sub-tasks such as tissue dissection offers a promising step toward reducing workload and expanding patient access. 
Prior work has relied on hand-crafted incision policies that cannot quantify uncertainty or has relied on simulation-based methods that require strong modeling assumptions. 
We instead view tissue attachment identification as an inherently probabilistic problem and propose a Bayesian approach that avoids explicit tissue modeling. 
Our method uses a Sequential Bayesian Hilbert Map (SBHM) to represent the likelihood that each tissue point is attached to the underlying resection surface. 
An ensemble of learned classifiers predicts attachment likelihoods from spatial data acquired during robotic tissue retraction, with each classifier serving as a noisy information source to update the SBHM. 
To plan the next retraction, we devise Bayesian Retraction Optimization (BRO) to select the most informative action under safety constraints. 
As the SBHM refines over time, regions with high attachment likelihood are selectively incised. 
We validate our method in simulation across diverse tissue geometries and acquisition strategies, and demonstrate zero-shot transfer to real robotic dissection experiments.

\end{abstract}

\section{Introduction}\label{sec:intro}
\input{intro}

\section{Related Work}
\input{related_work}

\input{problem_definition}

\input{method}

\section{Simulated Tissue Dissections}\label{sec:experiment}
\input{experiment}

\section{Real-World Application}\label{sec:real_experiment}
\input{real_experiment}

\section{Conclusions and Future Work}
\input{conclusion}

\appendices
\section{Bayesian Optimization Acquisition Functions}\label{sec:appendix}
\input{appendix}

\bibliographystyle{plain}
\bibliography{references}

\end{document}

%% file: intro.tex
\begin{figure*}
    \centering
    \includegraphics[width=0.9\linewidth, keepaspectratio]{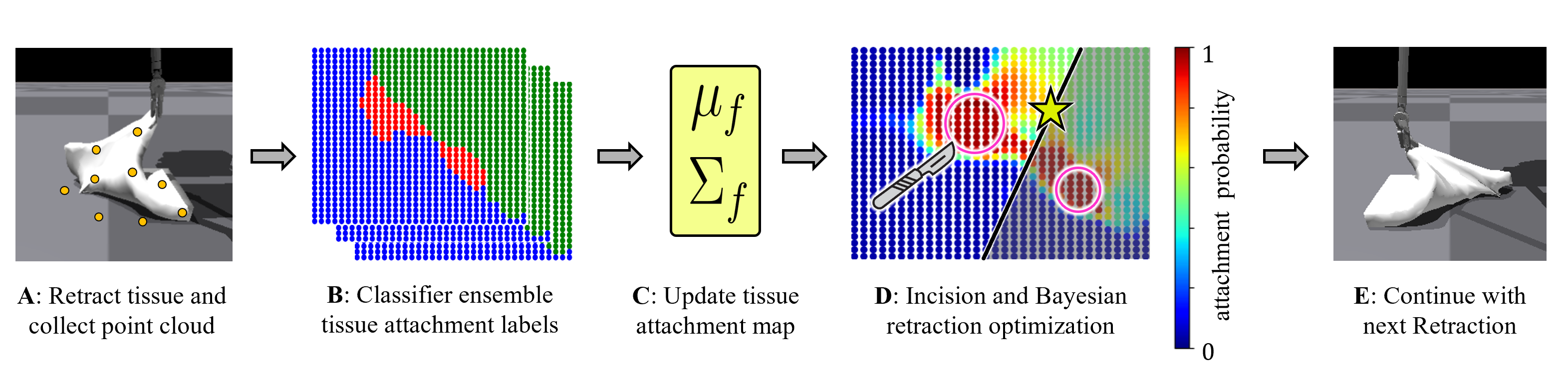}
    \vspace{-12pt}
    \caption{Illustration of a single iteration of our method. \textbf{A}: Robotic tissue retraction reveals surface geometry. \textbf{B}: A classifier ensemble predicts tissue attachment points from the point cloud (red: \textit{attached}, green: \textit{occluded}, blue: \textit{not attached}). \textbf{C}: Model parameters are updated to reflect all current knowledge about tissue attachment. \textbf{D}: The SBHM shows the probability that each point is attached to the resection surface. Highly likely attachment points (pink circles) can be dissected away. The next retraction is planned via Bayesian Retraction Optimization (BRO), selecting an acquisition point $q^*$ (gold star) that maximizes expected information gain while respecting tissue safety constraints. The shaded region indicates tissue planned to be lifted to expose the acquisition point.}
    \vspace{-12pt}
    \label{fig:method_overview}
\end{figure*}

As healthcare demand rises, the shortage of surgeons is expected to worsen over the next decade~\cite{zhang2020physician}. 
Reducing surgeon workload is therefore critical, and partial automation of surgical procedures offers a promising solution. 
Monotonous or time-consuming subtasks—such as retraction~\cite{nagy2018surgical, thach2024defgoalnet}, blunt dissection~\cite{nagy2018ontology}, suturing, and ablation~\cite{ostrander2024current, attanasio2021autonomy}—can be automated, freeing the surgeon to focus on higher difficulty or more critical tasks.

In this paper, we focus on surgical dissection, where a volume of tissue is removed from surrounding structures (e.g., tumor resection). 
Dissection is typically iterative: the surgeon (i) retracts the target tissue away from surrounding anatomy and (ii) cuts along the visible boundary. 
To decide where to cut next, surgeons actively manipulate tissue through a range of configurations, which serves two purposes. 
First, it exposes previously occluded regions for direct observation. 
Critically, retraction also reveals load-bearing connections, as tensioned or linearly stretched tissue indicates strong attachments and prime incision sites.

We formalize this strategy by predicting \textit{attachment points}, the regions along the tissue boundary where strong connections to surrounding structures indicate where incisions should be made. 
Identifying them is treated as an active sensing problem (e.g., active vision~\cite{chen2011active}). 
Prior to manipulation, all boundary points are considered potentially attached. 
The robot then deliberately retracts the tissue to induce informative deformations, revealing tensioned regions that provide evidence of attachment. 
After each retraction, the system updates its belief over attachment state based on the observed tissue. 
Evidence is accumulated over successive retractions, increasing confidence about where to cut.

To represent attachment state and uncertainty, we use a Sequential Bayesian Hilbert Map (SBHM)~\cite{senanayake2017bayesian}. 
SBHMs were originally developed in mobile robotics for continuous occupancy mapping using depth sensors such as LiDAR. 
In that setting, beam endpoints are labeled as occupied, and occupancy probability is updated sequentially over time. 
Here, we instead use this framework to model the probability that a boundary point is an attachment point.

During each retraction, endoscopic observations are processed by learned attachment classifiers that provide noisy binary labels. 
The outputs are incorporated into the SBHM, enabling principled fusion of uncertain evidence across viewpoints and over time. 
Probabilistic Hilbert map representations are known to be robust to noisy observations~\cite{ramos2016hilbert}. 
SBHMs in particular were designed to handle dynamic environments with uncertain updates~\cite{senanayake2017bayesian}. 
In contrast to prior work that considers noise in geometric range sensing, we address uncertainty arising from learned attachment classifiers.

To actively select informative tissue manipulations, we introduce Bayesian Retraction Optimization (BRO), which chooses retraction actions that maximize expected information gain about attachment state subject to safety constraints. 
When the SBHM indicates sufficient confidence, the robot performs an incision at the most probable attachment locations. 
This closed-loop cycle of retraction, observation, inference, and cutting continues until dissection is complete.

Prior work has explored automated identification of tissue attachment points for dissection~\cite{ge2021supervised, saeidi2019supervised, shinde2024jiggle}. 
However, existing approaches either use hand-crafted retraction strategies~\cite{saeidi2019supervised, ge2021supervised}, which cannot explicitly reason over uncertainty, or depend on deformable tissue simulations~\cite{shinde2024jiggle, boonvisut2014identification}. 
Simulation-based methods often require strong modeling assumptions (e.g., thin-shell approximations) and may suffer from a prohibitive sim-to-real gap. 
In contrast, our approach avoids deformable modeling assumptions and instead accumulates attachment evidence from classifiers.

Our contributions are: (i) a Bayesian framework for representing and sequentially updating tissue attachment maps without requiring an explicit tissue model, (ii) Bayesian Retraction Optimization (BRO), a method for generating highly informative retractions, (iii) simulated dissection experiments across diverse volumetric soft-tissue scenarios, and (iv) demonstration of zero-shot transfer in real-world tissue dissections using the da Vinci Research Kit (dVRK).  
Our results demonstrate that our approach can accurately localize attachment points without relying on deformable simulation, acquiring informative manipulations and outperforming two competitive baselines in simulation.
More broadly, this work highlights how active sensing strategies can be applied to surgical manipulation, showing that uncertainty-aware exploration can effectively guide complex surgical tasks.


%% file: related_work.tex
Several groups have developed methods for automating surgical retraction and incision.
Thach et al.\cite{thach2023deformernet} introduced a learning-based system that iteratively deforms tissue to match a predefined goal shape.
Saeidi et al.\cite{saeidi2019supervised} proposed a 3D path planning and control pipeline for electrocautery-based dissection, and Ge et al.\cite{ge2021supervised} built a supervised system for autonomous retraction and electrocautery.
Other approaches directly predict attachment points from a single view of the deformed tissue, as in Tagliabue et al.\cite{tagliabue2021data}.
While these methods demonstrate promising automation capabilities, they do not explicitly account for uncertainty in attachment point locations, which is an issue our framework is designed to address.

As discussed earlier, surgical dissection can benefit from active sensing by manipulating tissue to reveal hidden structures and reduce ambiguity about where to cut next.
An early example of this idea is Boonvisut et al.\cite{boonvisut2014identification}, which jointly tackled identification and exploration of boundary constraints using a simulator and a library of motion primitives. 
However, the expensive finite element simulations and exhaustive search limits clinical feasibility.
More recently, Shinde et al.\cite{shinde2024jiggle} proposed a model-based approach using a differentiable simulator and an Extended Kalman Filter to actively infer attachment points. 
While proven effective for thin-shell tissue models, their method makes simplifying assumptions that limit generalization to volumetric tissues and may suffer from real-to-sim gaps.
Other recent work has explored using Gaussian Processes (GPs) for active haptic sensing~\cite{yang2020efficient, garg2016tumor, salman2018trajectory}, but these methods have not yet been applied to attachment point identification.

In contrast, our approach explicitly models and reasons over uncertainty in attachment point locations, without relying on a tissue deformation model.
We represent attachment point probabilities using a Sequential Bayesian Hilbert Map (SBHM), analogous to occupancy mapping in mobile robotics.
Recent works have previously made this connection to formulate subsurface tumor localization as an occupancy mapping problem~\cite{garg2016tumor, salman2018trajectory, cho2021planning}.
Our work is the first to leverage a SBHM to model a probabilistic attachment point map for surgical dissection, and to optimize actions over that map.

%% file: problem_definition.tex
\section{Problem Definition}

At each discrete time step $t$, we carry out retraction, sensing, and (if confidence permits) incision. 
For this, we define $\mathcal{T}_t \subset \mathbb{R}^3$ to be the target deformable tissue, that we would like to remove via dissection.
The set $\mathcal{S} \subset \mathbb{R}^3$ is the attachment surface that $\mathcal{T}_t$ is attached to.
Each point in $\mathcal{S}$ is either an attachment point or is not, where attachment points are strong connections between $\mathcal{T}_t$ and $\mathcal{S}$ that we would like to remove by incision.

We partition the attachment surface into $\mathcal{S} = \mathcal{S}^{(1)}_t \cup \mathcal{S}^{(0)}_t$ where $\mathcal{S}^{(1)}_t$ is the set of attachment points and $\mathcal{S}^{(0)}_t$ are not attached.
We can say that $\mathcal{S}^{(1)}_t \subset \mathcal{T}_t$ by definition since tissue $\mathcal{T}_t$ is attached through these points. 
Note that while $\mathcal{S}$ does not change over time, the target tissue $\mathcal{T}_t$ and the attachment region $\mathcal{S}^{(1)}_t$ do change over time in general.

At any given time, the robot can grasp and retract the non-attached region $\mathcal{T}_t\setminus\mathcal{S}^{(1)}_t$.
Additionally, incision (i.e. cutting) can be carried out to remove attachment points from $\mathcal{S}^{(1)}_t$ to obtain $\mathcal{S}^{(1)}_{t+1}$. 
For sensing, we assume that we can only observe a partial-view point cloud $\mathcal{P}_t = C(\mathcal{T}_t)$ of tissue $\mathcal{T}_t$, where $C$ renders the tissue into a partial-view point cloud.
Thus, $\mathcal{S}^{(1)}_t$ may not be observable, but it can be estimated probabilistically by observing $\mathcal{P}_t$. At the subsequent time step, retraction, sensing and incision are informed by the prior knowledge gained in previous time steps. 

To reduce the search space, we discretize the attachment surface $\mathcal{S} = \mathcal{S}^{(1)}_t \cup \mathcal{S}^{(0)}_t$.
We consider a set of discrete candidate tissue attachment points chosen a priori that we call \textit{query points} $\mathcal{Q} = \{q_1,\cdots,q_N\}$ $\subseteq$ $\mathcal{S}$ within the surgical scene.
Since $\mathcal{S}$ is usually topologically equivalent to a 2D plane, in this paper, we choose $\mathcal{Q}$ to be a 2D grid of points representing the flat surface $\mathcal{S}$ that our tissue is fixed to.
Note that the selection of query points is arbitrary and could theoretically be a dense 3D grid in future work.
Similarly, define $y_t = [y_{t,1},\cdots,y_{t,N}]^T \in \{0,1\}^N$ to be the ground-truth attachment labels at time $t$.
The $j$ th query point $q_j$ is an attachment point at time $t$ if and only if $y_{t,j} = 1$. 
 
The overarching objective of the dissection task is to iteratively
\begin{enumerate}
    \item estimate the true attachment labels $y_t$ from partial-view point clouds $\{\mathcal{P}_i: i\leq t\}$ probabilistically, 
    \item execute incision to remove likely attachment points to form $y_{t+1}$,
    \item optimize robotic motions that retracts the tissue $\mathcal{T}_t$ into $\mathcal{T}_{t+1}$ with maximum information gain about $y_{t+1}$.
\end{enumerate}

%% file: method.tex
\section{Approach}

This overall method is illustrated in Figure~\ref{fig:method_overview} and detailed in Algorithm~\ref{alg:our_method}; each subroutine is explained in detail in the following sections.
The overall method is iterated until all tissue attachment points are removed via incision.
\begin{algorithm}
    \caption{Bayesian Dissection Automation}
    \label{alg:our_method}
    \textbf{input:} 
    Initial SBHM parameters $m_0$, $S_0$ and query points $\mathcal{Q}$.
    \\
    \textbf{initialize:}
    $q^* \in \mathcal{Q}$ at random,
    set
    $(m, S, \xi) \gets (m_0, S_0, 0)$
    \begin{algorithmic}
    \While{\texttt{tissue\_attached}($\mathcal{Q}$)}
        \State
        Robotic retraction to reveal $q^*$  \ref{subsec:acquisition}
        \State
        $\mathcal{P} \gets$ Partial point cloud observation
        \State
        $\mathcal{D} \gets \texttt{classification}(\mathcal{P}, \mathcal{Q})$ \ref{subsec:classifier}
        \State
        $m, S, \xi \gets \texttt{update}(m, S, \xi, \mathcal{D})$ \ref{subsec:updating}
        \State 
        $\mu_f, \Sigma_f \gets \texttt{predict}(m, S)$ \ref{sec:sbhm}
        \State
        Optional robotic incisions given $(\mu_f, \Sigma_f)$ \ref{subsec:incision}
        \State 
        $q^* \gets 
        \underset{q \in \mathcal{Q}}{\arg\max} \,\,\, 
        \texttt{acquisition}(q, \mu_f, \Sigma_f)$ \ref{subsec:acquisition}, \ref{sec:appendix}
        \EndWhile
    \end{algorithmic}
\end{algorithm}
As tissue is retracted, point cloud data is collected and processed by an attachment point classifier, which acts as a ``noisy sensor'' to iteratively update the SBHM in a principled way. 
This approach provides a probabilistic understanding of tissue attachment that improves over time. 
Because different retractions reveal different information, we use BRO to select the most informative retraction to command. 
Once the SBHM indicates sufficient confidence that a tissue location is an attachment point, it is removed via incision.

\subsection{Sequential Bayesian Hilbert Map}
\label{sec:sbhm}

Throughout dissection, we model each query point as having a binary attachment state $y \in \{0, 1\}$, where $y = 1$ indicates that the point is attached. 
The SBHM represents the predictive distribution $p(y = 1 \mid q)$ for all $q \in \mathcal{Q}$, integrating all past observations into a global map of attachment likelihoods. 

In particular, the current state of the SBHM is encoded in the mean and covariance parameters $(m, S)$ which specify a Gaussian distribution $w \in \mathbb{R}^{d} \sim \mathcal{N}(m, S)$.
Given $w$, we can compute the conditional sigmoid likelihood 
\begin{equation}
    p(y=1 \mid q, w) = \sigma(w^T\phi(q)),
\end{equation}
where $\phi(q) = \begin{bmatrix} 1, k(q,\hat{x}_1), \cdots, k(q,\hat{x}_M) \end{bmatrix}^T$ is a feature vector, the $M$ points $\hat{x}_i \in \mathbb{R}^3$ are fixed hinge points, and $k$ is a kernel function, which we take as the radial basis function kernel with bandwidth $\gamma$.

Given the parameters $(m, S)$, the predictive distribution can be approximated as follows~\cite{bishop2006pattern}: 
\begin{equation}
\begin{gathered}
    \label{eq:mean_approx}
    p(y=1 \mid q, m, S) = \int p(y=1 \mid q, w) p(w \mid m,S) dw \\
    \approx \sigma 
    \left (
        \frac
        {m^T \phi(q)}
        {(1 + \pi (\phi(q)^T S \phi(q))^2/8)^{1/2}} 
    \right ) 
    .
\end{gathered}
\end{equation}
Thus, we can compute the likelihood that each point in $\mathcal{Q}$ is an attachment point:
\begin{equation}
    \label{eq:f_w}
    f(w) = 
    \begin{bmatrix}
        p(y = 1|q_1, m, S),
        \dots,
        p(y = 1|q_N, m, S)
    \end{bmatrix}^T
    ,
\end{equation}
which we approximate as a Gaussian: $f(w) \sim \mathcal{N}(\mu_f, \Sigma_f)$.
The mean $\mu_f$ is approximated using (\ref{eq:mean_approx}), and the covariance $\Sigma_f$ is estimated as the sample covariance of outputs obtained by Monte Carlo sampling $w$ and propagating through (\ref{eq:f_w}).

Overall, the parameters $(m, S)$ compactly encode the current belief about tissue attachment across all query points in the map.
In the following sections, we describe how these parameters are iteratively updated using noisy classifier measurements as new data become available.

\subsection{Tissue Attachment Point Classifier}
\label{subsec:classifier}

During each robotic retraction, a partial-view point cloud $\mathcal{P}$ of the deformed tissue is collected to update our estimate of tissue attachment state (see Figure~\ref{fig:method_overview}-A).
Because the classifiers are deployed zero-shot after being trained entirely in simulation, robustness to noisy and out-of-distribution observations is critical.
To address this, we use an ensemble of PointNet-based classifiers~\cite{qi2017pointnet}, which reduces sensitivity to out of distribution point clouds.
Each classifier $f_{\theta_j}$ predicts the attachment state for each query point $q \in \mathcal{Q}$, conditioned on $\mathcal{P}$ (see Figure~\ref{fig:method_overview}-B), producing a categorical distribution $\mu_{\theta_j}(q, \mathcal{P}) \in \Delta^3$ over three classes: not attached (0), attached (1), and occluded (2).
The ensemble is denoted $\{ f_{\theta_j} \}_{j=1}^{N_e}$, and training details are provided in Section~\ref{sec:training_data}.

To obtain a robust occlusion mask, we average the predicted occlusion probabilities across the ensemble. 
Points with mean occlusion probability greater than $0.5$ are treated as occluded and filtered out. 
The remaining, non-occluded measurements form the dataset
\begin{equation}
\begin{gathered}
    \mathcal{D} = 
    \{ (q, y) \mid 
    q \in \mathcal{Q},\
    y = f_{\theta_j}(q, \mathcal{P}),\ 
    j = 1, \dots, N_e
    \}
    \\ \text{s.t.} \quad
    \tfrac{1}{N_e} \sum_{j=1}^{N_e} (\mu_{\theta_j}(q, \mathcal{P}))_3 \le 0.5.
\end{gathered}
\end{equation}
At test time, each classifier prediction $f_{\theta_j}(q, \mathcal{P})$ corresponds to the class with the highest predicted probability in $\mu_{\theta_j}(q, \mathcal{P})$.
The filtered dataset $\mathcal{D}$ is used to update the SBHM at the current timestep (Algorithm~\ref{alg:our_method}).

\subsection{Updating the SBHM with Classifier Measurements}
\label{subsec:updating}

Here we explain how the dataset $D = \{(q_i, y_i)\}^{N_D}_{i = 1}$ is used to update SBHM (see Figure \ref{fig:method_overview}-C). 
Given the prior $\mathcal{N}(w|m_{t - 1}, S_{t - 1})$, the posterior $\mathcal{N}(w|m_{t}, S_{t})$ is computed using Variational Bayesian Logistic regression~\cite{jaakkola1997variational,bishop2006pattern}, which manifests as an Expectation Maximization (EM) Algorithm.
The expectation step of EM updates the posterior:
\begin{equation}
\begin{split}
    m^\text{new} &= 
    S^\text{new}
    \left \{
        (S^\text{old})^{-1}m^\text{old} + 
        \sum_{i = 1}^{N_D}
        \left (
            y_i - \frac{1}{2}
        \right )
        \phi(q_i)
    \right \}
    , \\
    (S^\text{new})^{-1} &= 
    (S^\text{old})^{-1} + 
    2\sum_{i = 1}^{N_D} \lambda(\xi_i^\text{old})\phi(q_i)\phi(q_i)^T
    ,
\end{split}
\end{equation}
where $\lambda(\xi_i) = \frac{1}{2\xi_i}[\sigma(\xi_i) - \frac{1}{2}]$. 
The $\xi_i$ are a set of variational parameters that are updated to maximize the variational lower bound during the subsequent maximization step of EM:
\begin{equation}
    \xi^\text{new}_i  =
    \sqrt{\phi(q_i)^T(S^\text{new} + m^\text{new}(m^\text{new})^T)\phi(q_i)}
    .
\end{equation}

To carry out EM, first  $m^\text{old}$ and $S^\text{old}$ are initialized to $m_{t - 1}$ and $S_{t - 1}$, and the variational parameters $\xi = (\xi_1,...,\xi_{N_D})$ are initialized to $0$. 
Note that during the first iteration, we initialize $m_0 = 0$ and $S_0 = \sigma I$ where $\sigma$ is very large, indicating essentially zero prior information about tissue attachment points before dissection.
Next, the expectation and maximization steps are executed iteratively to convergence.
The final values of $m^\text{new}$ and $S^\text{new}$ are taken as the updated parameters $m_{t}$ and $S_{t}$, representing the new SBHM that encodes all available tissue attachment information.

\subsection{Robotic Attachment Point Incision}
\label{subsec:incision}

As additional classifier data arrive, the SBHM gradually becomes more confident about which points are truly attached, ready to be removed (i.e. incision candidates).
Our focus is on probabilistic reasoning rather than motion planning; thus, we assume the existence of a mechanism capable of detaching any chosen set of points.
In simulation, we can simply delete the corresponding attachments between surfaces.
To determine which points should be removed, we classify a non-occluded query point $q \in \mathcal{Q}$ as an incision candidate if at least half of its $K = 15$ nearest neighbors have predictive probabilities $p(y = 1 \mid x)$ above the incision confidence threshold of $0.9$.
This neighborhood-based rule enforces spatial consistency and mitigates the effect of outliers.

In our real-world experiments (Section~\ref{sec:real_experiment}), classifier noise increases uncertainty in the SBHM, as the method is applied zero-shot after being trained entirely in simulation.
To account for this, we lower the incision confidence threshold to $0.83$ and relax the neighborhood rule: a query point is considered an incision candidate if at least $41\%$ of its $K = 15$ nearest neighbors exceed the adjusted threshold. These hyperparameters are chosen manually based on real-world point clouds to trade off false positives and false negatives.

\subsection{Bayesian Retraction Optimization}
\label{subsec:acquisition}

\begin{figure}
  \centering
  \includegraphics[width=1.0\linewidth, trim=0pt 10pt 0pt 11pt, clip]{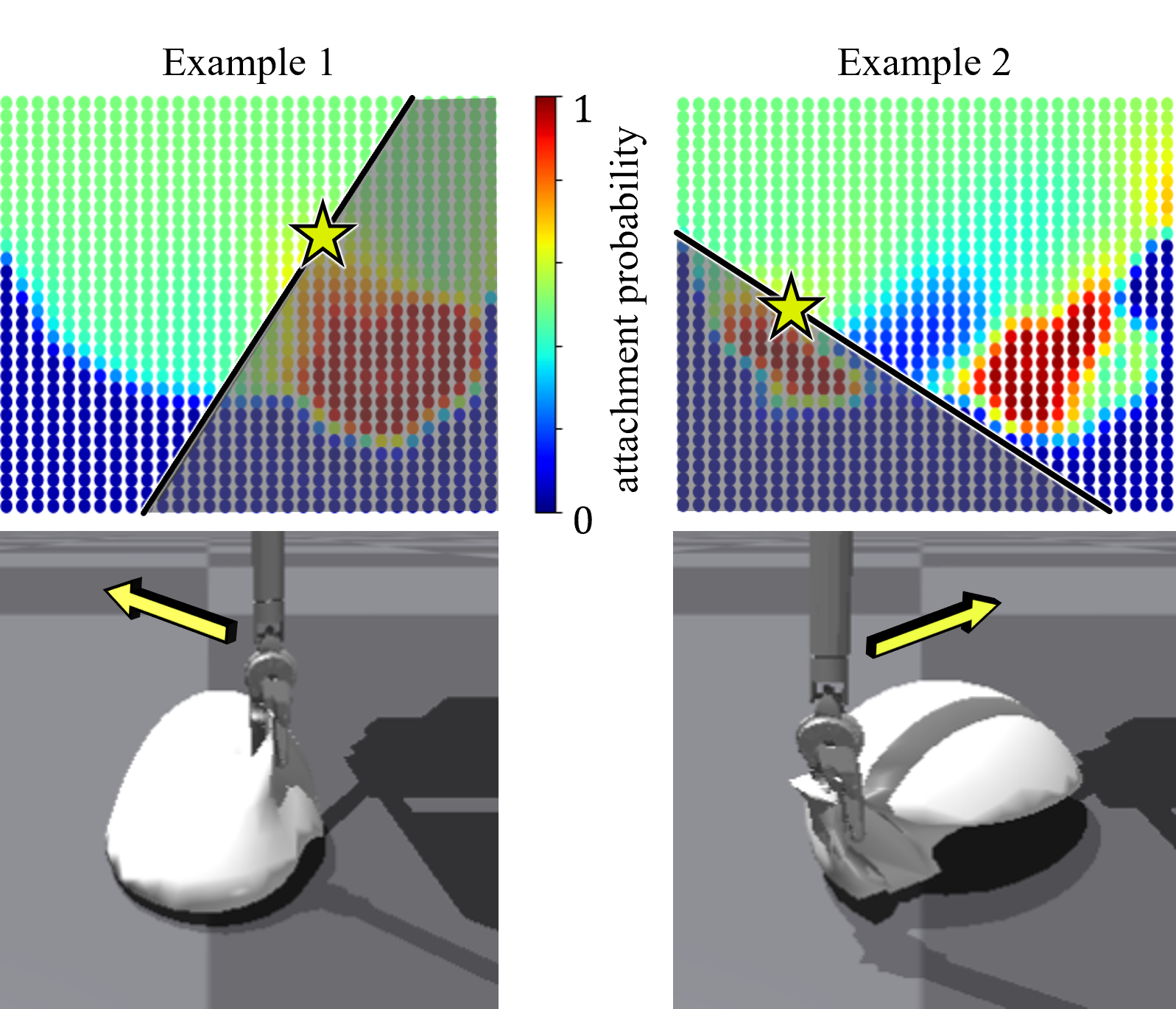}
  \caption{Bayesian retraction optimization examples from our simulations. Given the current SBHM (heat maps), an acquisition point $q^*$ (gold star) is selected to reveal to the camera which maximizes expected information gain (i.e. active sensing).
  The actual retraction motion is planned by a separate optimization, which selects a tissue region containing  $q^*$ to lift vertically (shaded region of map) which minimizes total tissue tearing probability.}
  \label{fig:retractions}
  \vspace{-10pt}
\end{figure}

We aim for robotic retraction to reveal \textit{complementary} information relative to the current tissue attachment state in the SBHM. 
For simplicity, we choose a retraction that exposes a specific query point $q^* \in \mathcal{Q}$ to the camera.
The acquisition point $q^*$ is selected to maximize an acquisition function, which quantifies how informative a data point is likely to be. 
In our simulations, we compare Expected Improvement (EI)~\cite{jones1998efficient}, noisy Expected Improvement (nEI)~\cite{zhou2024corrected}, and the Upper Confidence Bound (UCB)~\cite{srinivas2009gaussian} as acquisition functions (Appendix \ref{sec:appendix} details their evaluation).
The optimal $q^*$ is obtained by maximizing the acquisition function (i.e., Bayesian optimization) via exhaustive search over $\mathcal{Q}$.

Given the acquisition point $q^*$, it is still nontrivial to command a specific robot gripper motion to reveal $q^*$ to the camera.
Therefore, we leverage DeformerNet~\cite{thach2023deformernet}, a shape servoing method for deformable object manipulation. 
Given a desired tissue goal shape and the current tissue point cloud, DeformerNet computes an end effector trajectory to achieve the goal. 
To lift the tissue and expose $q^*$, we heuristically define the goal shape using a vertical plane $ax_1 + bx_2 = 0$ orthogonal to the planar attachment surface (see \cite{thach2023deformernet}). 
Tissue points on one side of the plane, $\{x \in \mathcal{P}: ax_1 + bx_2 \ge 0\}$, are rotated up by 90 degrees to reveal the query points beneath, while points on the other side remain unchanged. 
The vertical plane parameters are chosen by solving:
\begin{equation}
    \underset{a,b \in \mathbb{R}}{\text{minimize}} \sum_{x \in Q_{0}}
    p(y=1|x, m, S)
    \quad
    \text{s.t.} \quad
    x^* \in Q_0,
\end{equation}
where the objective penalizes retractions that lift attached tissue and $Q_0=\{x \in \mathcal{Q}: ax_1 + bx_2 \ge 0\}$. 
Currently, this problem is solved via random sampling. 

Example retractions generated using this approach in our simulations are shown in Figure~\ref{fig:retractions}.
Our vertical plane heuristic for revealing the optimal $q^*$ is simplified by assuming a planar attachment surface.
More expressive tissue goal shape generation methods \cite{thach2024defgoalnet} could generalize this process to arbitrary attachment surface geometries in future work.

%% file: experiment.tex
\begin{figure}
  \centering
  \includegraphics[width=1.0\linewidth]{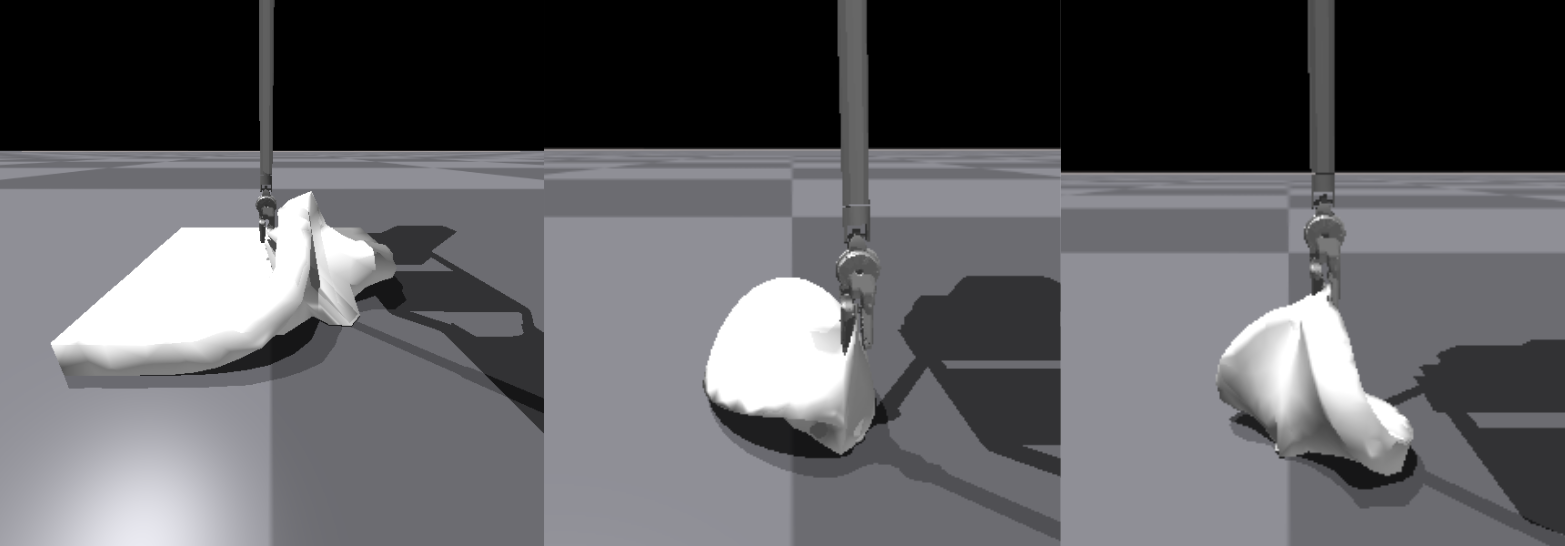}
  \caption{Tissue dissection simulation environment for evaluating our approach. We use the Isaac Gym simulator with the dVRK system to test on several tissue geometries: \textit{Left}—BOX, \textit{Center}—ELLIPSOID, \textit{Right}—CYLINDER.}
  \label{fig:simulation_scenes}
\end{figure}

For both classifier training and our simulated dissection procedures, we simulate tissue interaction with the da Vinci Research Kit (dVRK)~\cite{kazanzides2014open} patient-side manipulator in the Isaac Gym simulation environment \cite{liang2018gpu}.
Point cloud data is acquired via a simulated RGBD camera at a fixed pose.
We ran our simulations using three geometric primitive classes representing tissue volumetrically, which we call BOX, CYLINDER, and ELLIPSOID (see Figure \ref{fig:simulation_scenes}).
The tissue parameter statistics used in our simulations are summarized in Table \ref{tab:tissue_specs}.
\begin{table}
    \caption{Specification of tissue parameter statistics used in BOX, CYLINDER, and ELLIPSOID simulations.}
    \label{tab:tissue_specs}
    \centering
    \begin{tabular}{ccc}
        \toprule
        \textbf{\shortstack{Geometric\\Primitive}} & \textbf{\shortstack{Young's\\Modulus (Pa)}} & \textbf{\shortstack{Number of\\Attachment Discs}} \\
        \midrule
        BOX            & $\mathcal{N}(1k,\, 0.2^2k)$ & $\mathcal{U}(1,\,10)$ \\
        CYLINDER       & $\mathcal{N}(1k,\, 0.2^2k)$ & $\mathcal{U}(1,\,8)$ \\
        ELLIPSOID & $\mathcal{N}(5k,\, 1k)$     & $\mathcal{U}(1,\,3)$ \\
        \bottomrule
    \end{tabular}
    \\
    \vspace{2mm}
    \parbox{\columnwidth}{\footnotesize 
    \textit{Note.} The length and thickness of each tissue are drawn from the uniform distributions 
    $\mathcal{U}(7.5\,\mathrm{cm},\,15\,\mathrm{cm})$ and $\mathcal{U}(1\,\mathrm{cm},\,1.5\,\mathrm{cm})$, respectively. 
    Length refers to diameter for the CYLINDER and ELLIPSOID. 
    The ELLIPSOID is constructed by narrowing one axis of a hemisphere using a random aspect ratio. 
    For BOX, CYLINDER, and ELLIPSOID, the top 15, 15, and 45 query points closest to the center of each attachment disc are labeled as attachment points, respectively.}
    \vspace{-10pt}
\end{table}

\subsection{Classifier Training}
\label{sec:training_data}

To collect a single training example, we first generate a random tissue by sampling using the statistics in Table \ref{tab:tissue_specs}.
The tissue is then randomly attached to the 2D resection surface by sampling a random configuration of attachment discs which are circular regions of attachment between the simulated tissue and resection surface.

To simulate retraction, we generate a random retraction by sampling random vertical plane parameters and executing retraction using the method in Section \ref{subsec:acquisition}.
To generate the ground-truth labels, we project a ray from the camera to each query point. 
If the ray intersects the tissue mesh, then the query point is labeled occluded.
For each attachment disk, if one of its attachment points is not occluded, then all the attachment points of the disk are labeled as attached.
The points that are neither attached nor occluded are labeled as not attached.

We collected 11,881 BOX, 4,407 CYLINDER, and 10,157 ELLIPSOID examples for classifier training. 
An ensemble of 20 classifiers, each with a different seed, was independently trained for 160 epochs each in PyTorch using Adam with a learning rate scheduler decaying by 0.1 every 80 epochs. 
To address class imbalance, the attached class (1) was upweighted by a factor of 6 in the cross-entropy loss. 
The test set contained 401 BOX, 200 CYLINDER, and 200 ELLIPSOID examples. 
Averaging across the ensemble, the attached-class precision and recall were 0.343 and 0.584, respectively.


\subsection{Simulated Dissection Experiments}

\begin{figure}
    \centering
    \includegraphics[width=1.0\columnwidth, trim=10pt 0pt 2pt 11pt, clip]{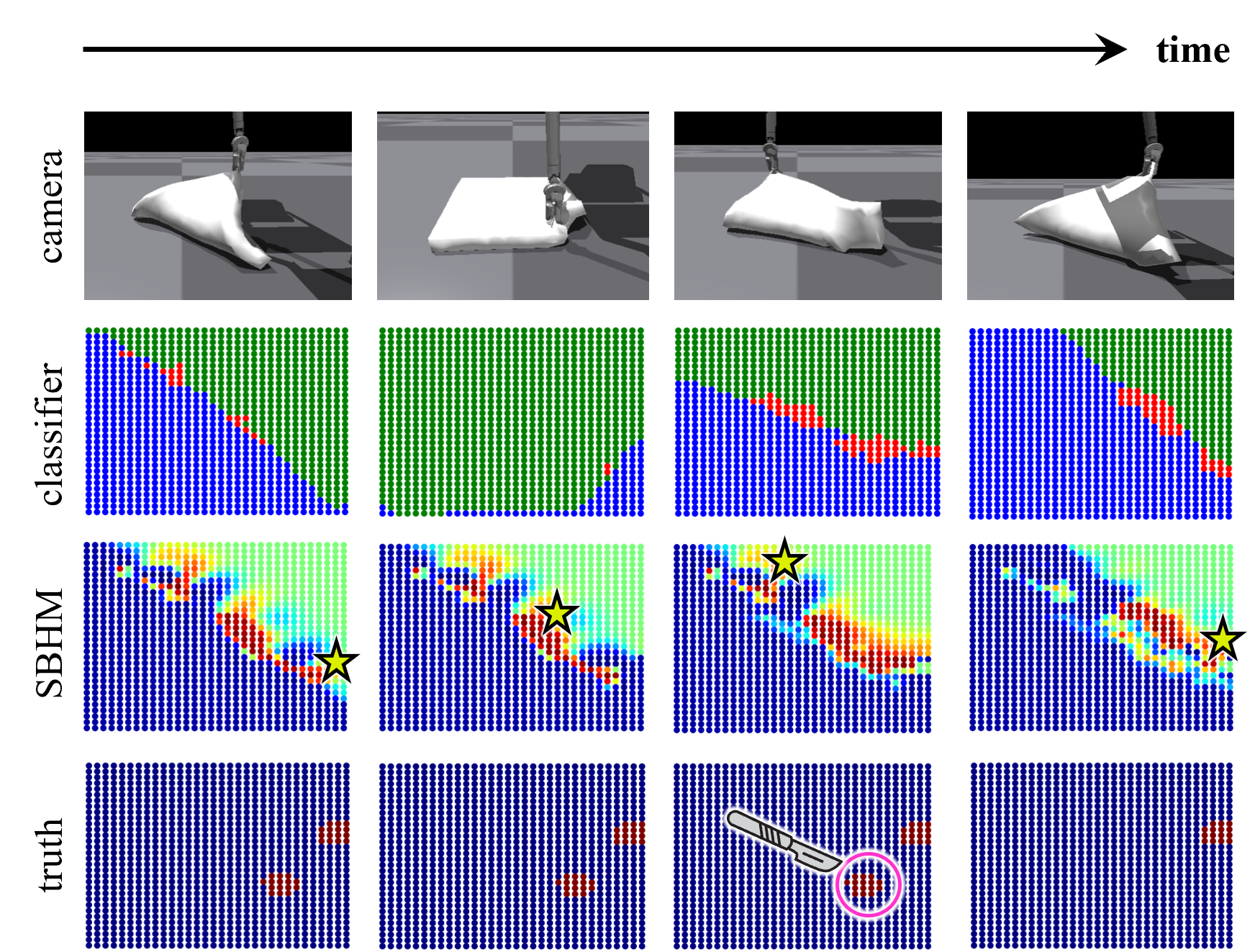}
    \\ \vspace{5pt} 
    \includegraphics[width=0.75\columnwidth, trim=10pt 0pt 2pt 0pt, clip]{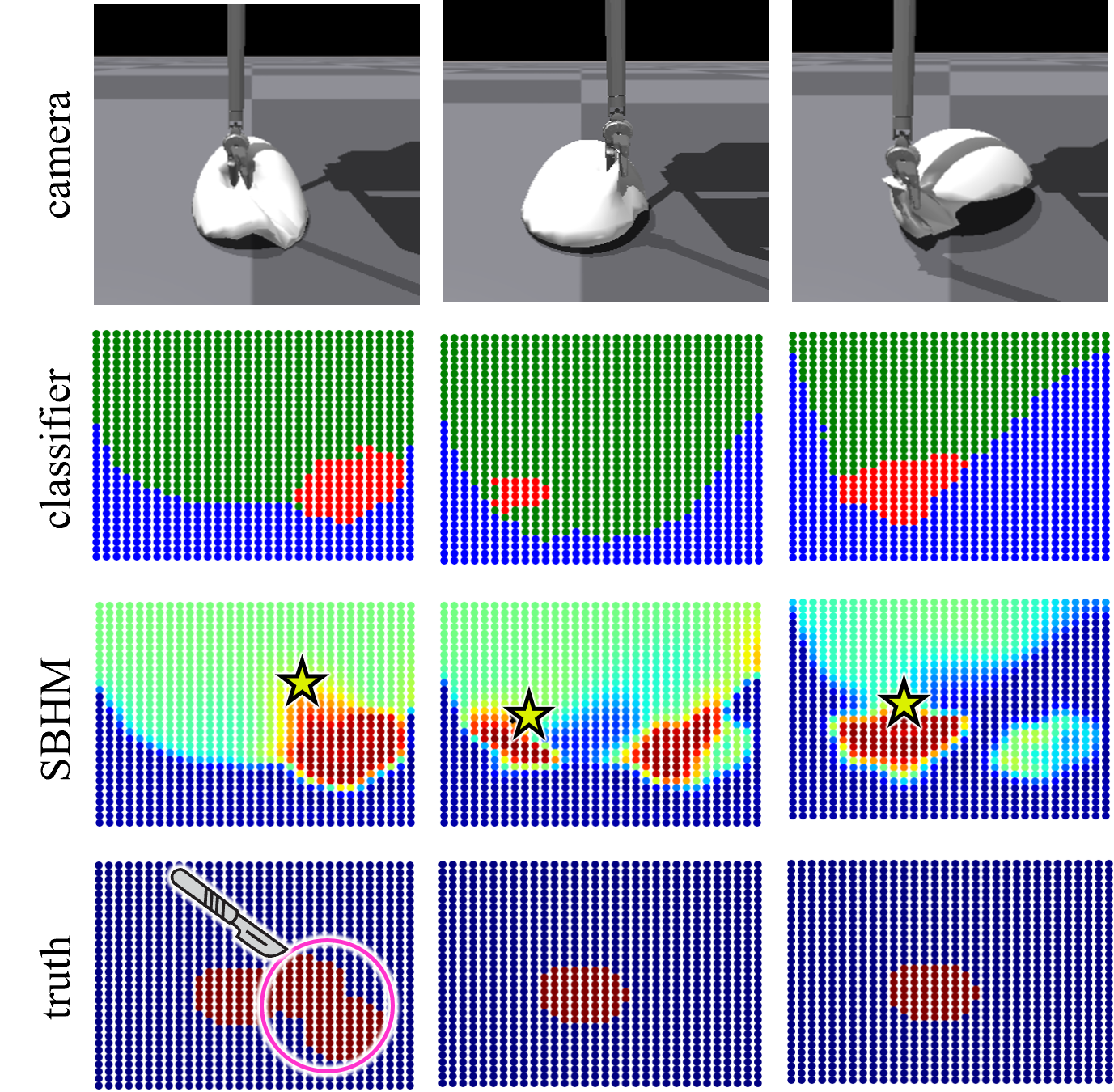}
    \\ \vspace{5pt}
    \includegraphics[width=1.0\columnwidth, trim=10pt 0pt 2pt 0pt, clip]{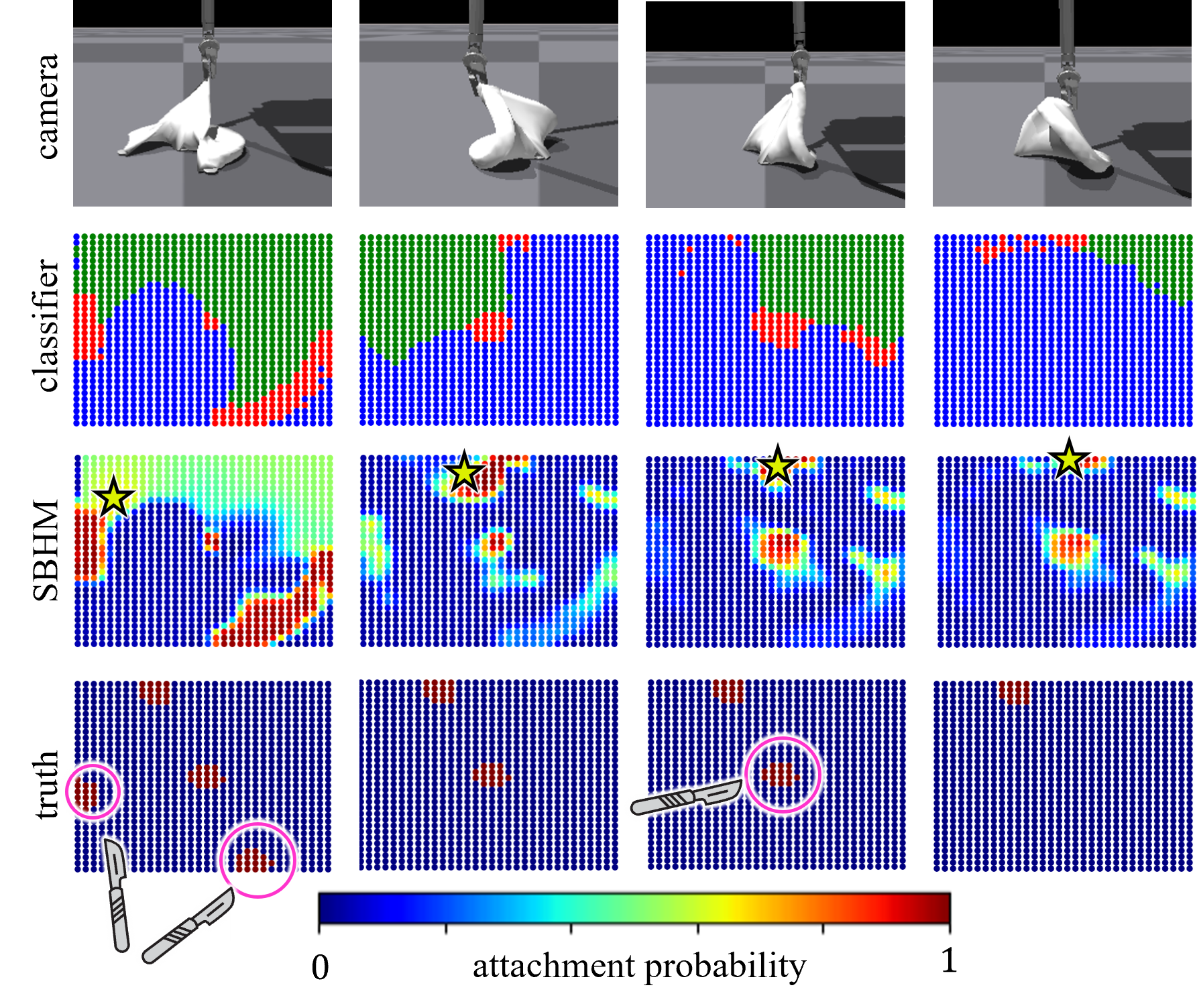}
    \caption{Example tissue dissection execution sequences in our simulations for each geometric primitive: \textit{top}: BOX, \textit{middle}: ELLIPSOID, and \textit{bottom}: CYLINDER. At each time step, our method leverages the SBHM to decide whether to remove highly certain attachment points (pink circles) and to optimize the next acquisition point for robotic retraction (gold star). Note we show a single classifier from the ensemble for clarity.}
    \label{fig:execution_sequences}
\end{figure}

We evaluated BRO in numerous simulated dissection tasks across various tissue geometries. 
The simulation setup mirrors the training environment, and Algorithm~\ref{alg:our_method} is implemented in Python using Isaac Gym~\cite{liang2018gpu}. Tissue incision is modeled by removing attachment discs corresponding to predicted incision points (Section~\ref{subsec:incision}).
To account for spatial mismatch between candidate points and discs, we expand each disc radius by 5 mm when reporting results.
The process runs for 20 time steps or until all attachment discs are removed.
SBHM hyperparameters are set as follows: $\gamma = 10{,}000$ (Section~\ref{sec:sbhm}), with 20 EM iterations per update and hinge points coinciding with query points. 
Acquisition function parameters for EI, nEI, and UCB (Appendix~\ref{sec:appendix}) are 1.0, 1.0, and 2.0, respectively.

For each geometric primitive (BOX, CYLINDER, ELLIPSOID), we simulated 30 random trial runs of the dissection task. 
Figure~\ref{fig:execution_sequences}, shows an example execution sequence for each geometric primitive.

\subsection{Simulation Results}
 
\begin{figure}
    \centering
    \includegraphics[width=1\linewidth, trim=10pt 10pt 0pt 10pt, clip]{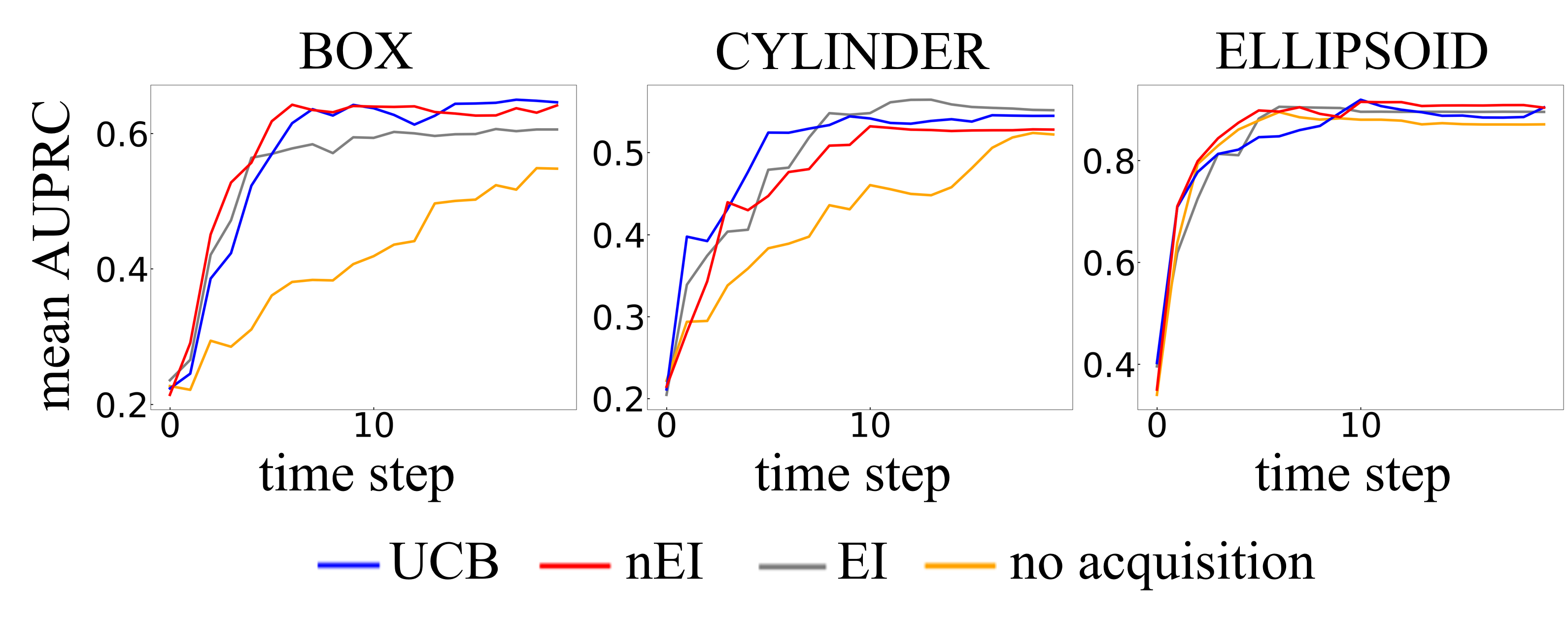}
    \caption{Average precision-recall area (AUPRC) over all trials at each simulation time step. Regardless of acquisition function, BRO significantly outperforms random (``no acquisition'') retractions for identifying tissue attachment points.}    \label{fig:comparing_acquisition}
\end{figure}

We evaluate the performance using the Area Under Precision-Recall Curve (AUPRC) at each timestep averaged over all trials, which compares the predictive mean \ref{eq:mean_approx} against ground-truth attachment labels. 
If there are no attachment disks left at certain timestep, then AUPRC is 1.
To demonstrate the effectiveness of retraction optimization, we additionally compare our method to a baseline ``no acquisition'', which means that the $q^*$ is selected uniformly at random from $\mathcal{Q}$. 
As shown in Figure \ref{fig:comparing_acquisition}, BRO significantly outperforms the baseline in general.
\begin{figure}
    \centering
    \includegraphics[width=1\columnwidth, trim=10pt 10pt 0pt 10pt, clip]{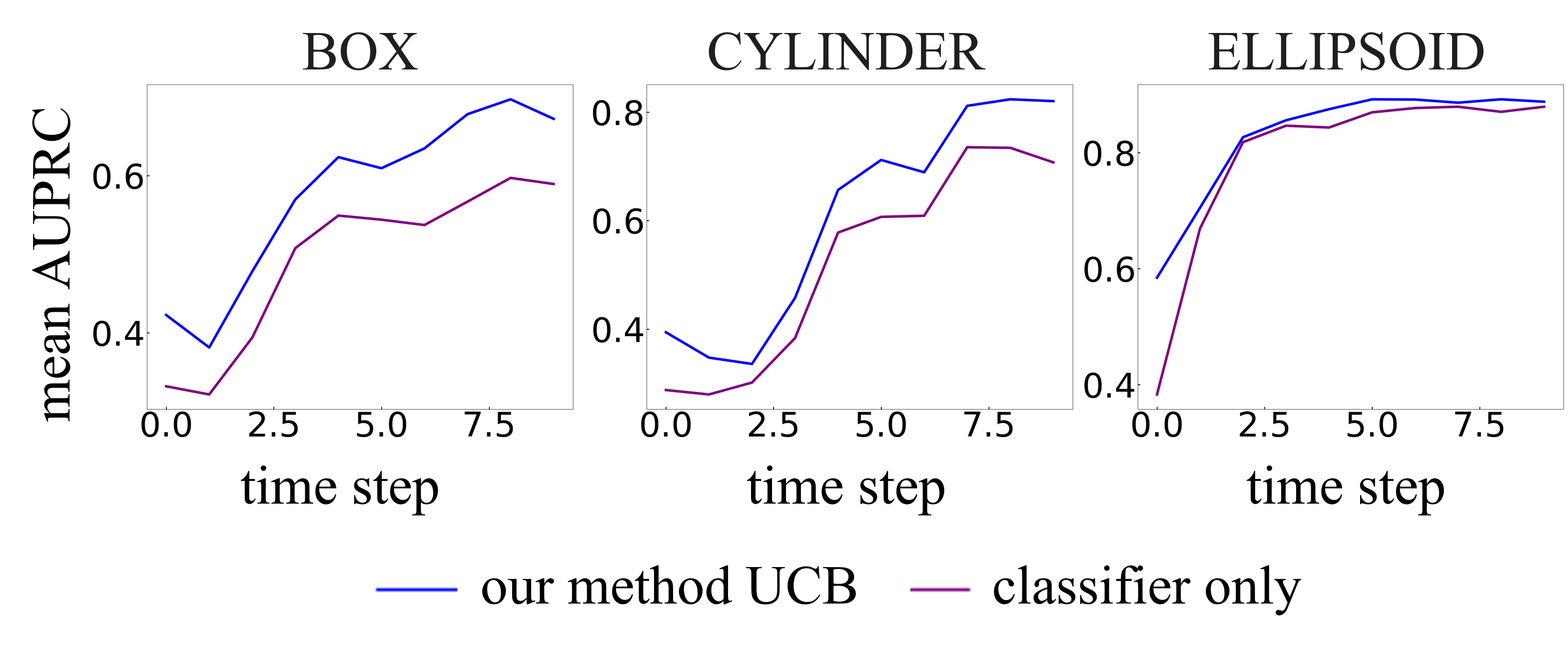}
    \caption{Average performance (AUPRC) of our method with simulated classification errors. Using an SBHM to represent and update attachment probabilities significantly improves identification performance, compared with the output of a single classifier.}
    \label{fig:classifier_comparison}
    \vspace{-10pt}
\end{figure}

We next evaluate the benefit of using an SBHM under classifier noise, which can be substantial with real data. 
For this experiment, the SBHM is updated using a single noisy classifier, with Gaussian noise (std. dev. of 2) added to the classifier logits before each update. 
We run 30 simulated dissections, performing incisions and retractions according to the SBHM predictive distribution as normal.
The resulting SBHM maps are compared to the raw classifier outputs at each time step over non-occluded points. 
As shown in Figure~\ref{fig:classifier_comparison}, reasoning with the SBHM yields a substantial performance gain.
 
Next we evaluate performance in reasoning over the SBHM for incision decisions (Section~\ref{subsec:incision}). 
A candidate incision point is correct if it overlaps with an attachment disk, and incision accuracy is defined as the fraction of candidate points that are correct. 
Across our simulations, the average incision accuracies for BOX, CYLINDER, and ELLIPSOID were 0.549, 0.681, and 0.760, respectively.

%% file: real_experiment.tex


\begin{figure*}
  \centering
  \includegraphics[width=0.9\textwidth]{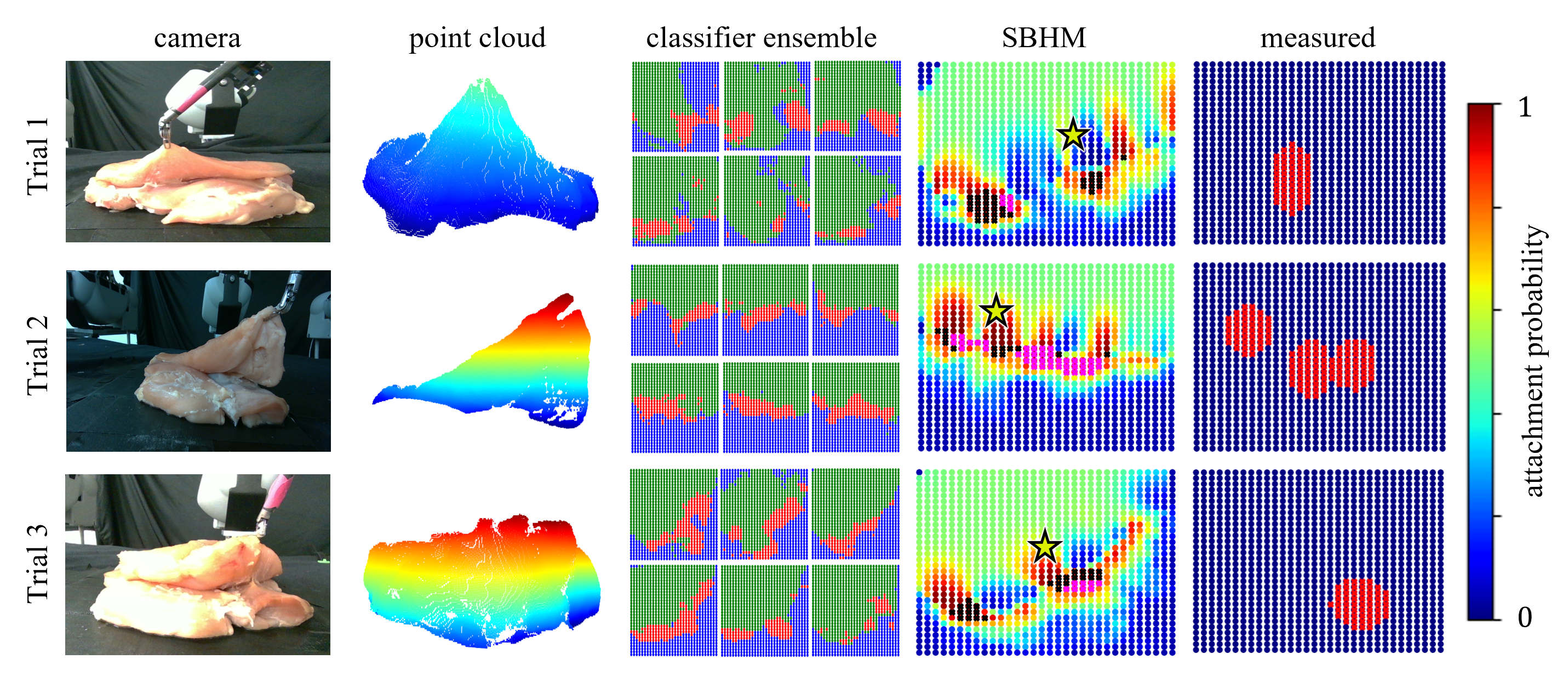}
  \caption{Experimental results with the dVRK system in chicken breast tissue showing identification of attachment points after a single iteration of our method. Across all three trials, the predictive distribution is a reasonable match with the measured tissue attachment locations. The pink and black points denote correct and incorrect identification of tissue attachment points respectively, and the gold star indicates the next acquisition point.}
  \label{fig:single_step_results}
\end{figure*}

\begin{figure*}
  \centering
  \includegraphics[width=0.9\textwidth]{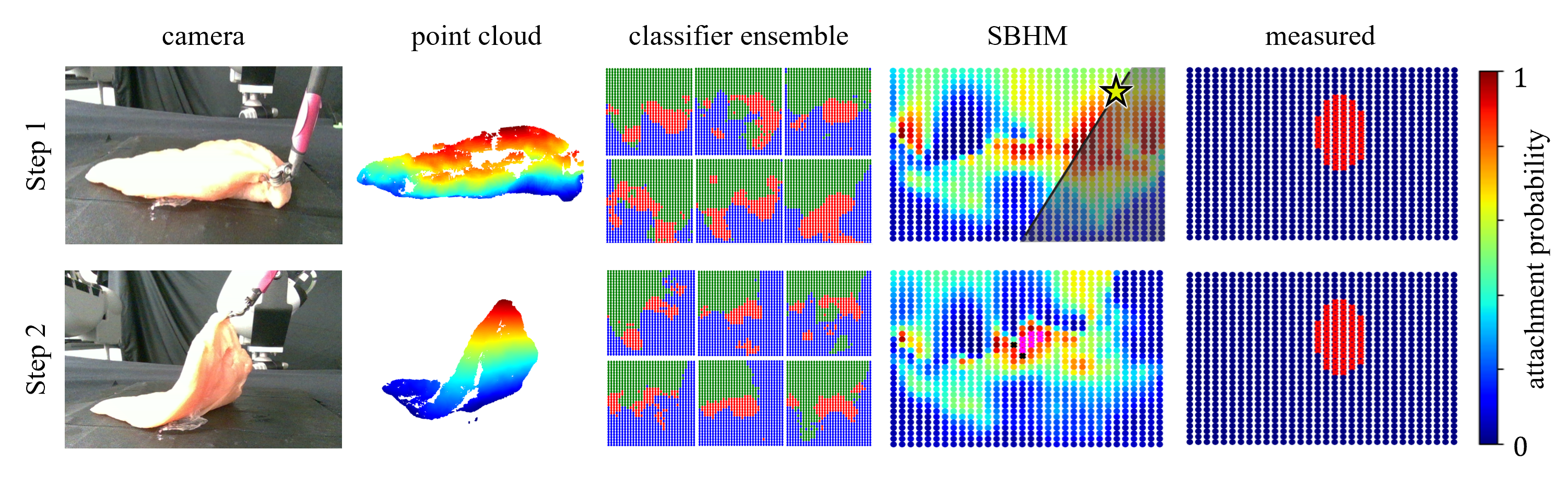}
  \caption{Example of a multi-timestep tissue dissection using our method. Starting from the initial SBHM (first row), the selected acquisition point (gold star) produces an informative retraction (second row, camera view), resulting in an improved SBHM update. Pink and black points denote correct and incorrect identification of tissue attachment points compared with the measured locations. Cutting these points would likely yield an effective incision in this case.}
  \label{fig:multi_step_results}
\end{figure*}

We transfer our method \textit{zero-shot} to real-world surgical dissection using ex vivo chicken breast tissue and the da Vinci Research Kit (dVRK) for robotic retraction.
We use an Intel RealSense D405 depth camera to capture tissue point clouds.
As in simulation, tissues are attached to a planar resection surface using multiple discrete attachment disks, bonded to the surface with cyanoacrylate adhesive.
The RGB images were segmented with Segment Anything~\cite{kirillov2023segment}, and the resulting masks are applied to segment the point cloud data.
We additionally record the locations of the attachment disks and transform it into the coordinate frame of the query points for comparison with our method.
We conducted four dissection experiments in total.
In three cases, all attachment disks were correctly identified in a single iteration (Figure~\ref{fig:single_step_results}); in one case, two retractions were required to confidently identify all attachment regions (Figure~\ref{fig:multi_step_results}).

Overall, these experiments presented two key challenges:
(1) real RGB-D point clouds are noisy, making surface deformation difficult to perceive even for a human observer, and
(2) the real tissue properties differ significantly from the simulated ones.
As a result, the real point clouds are out of the training distribution of the attachment classifiers, leading to extremely noisy attachment predictions and motivating our robustness strategies in Sections~\ref{subsec:classifier} and~\ref{subsec:incision}.

Despite these challenges, our method correctly predicted tissue attachment points in all cases. Despite false-positive predictions in Figure~\ref{fig:single_step_results}, they are in fact visually plausible attachment points based on the point clouds. 
Furthermore, in the multi-step dissection (Figure~\ref{fig:multi_step_results}), incision predictions became progressively more accurate as uncertain regions were revealed, demonstrating the effectiveness of our sequential Bayesian approach.

%% file: conclusion.tex
We have presented a new framework for probabilistic tissue attachment mapping during surgical dissection, demonstrating good performance across a range of simulated dissections.
Our results show that BRO substantially outperforms random acquisition, and that the SBHM maintains mapping accuracy even under classifier noise, highlighting the importance of sequential probabilistic reasoning for out-of-distribution data.
Our real-world experiments further demonstrate zero-shot sim-to-real transfer, successfully identifying tissue attachment regions.

While planar attachment surfaces are a useful preliminary benchmark, future work will consider more real-world experiments with non-planar and anatomically realistic tissue geometries typically encountered in tumor resection.
This will require more general methods for generating tissue goal shapes that reveal the desired acquisition point to the camera.
Additionally, more expressive classifier models (e.g. diffusion models) and improved domain adaptation techniques (e.g. transfer or self-supervised learning) may be necessary for robust real-world performance.
Finally, we aim to integrate the SBHM directly into probabilistic incision planning, enabling the robot to reason about uncertainty in tissue attachment when selecting and executing cutting motions.

%% file: appendix.tex
Here we list the Acquisition functions that we use for BRO to select highly informative retraction motions.
Each assumes knowledge of the predictive distribution (i.e. $\mu_f$ and $\Sigma_f$), computed based on the current SBHM parameters $(m,S)$ (Section \ref{sec:sbhm}).
In this paper, we use the Expected Improvement (EI)~\cite{jones1998efficient}, noisy Expected Improvement (nEI)~\cite{zhou2024corrected} and the Upper Confidence Bound (UCB)~\cite{srinivas2009gaussian} as acquisition functions for BRO.
First and foremost,
\begin{equation}
\begin{gathered}
    \text{EI}(q) = 
    (\mu_f(q) - \mu_f(q^+) - \epsilon)
    \Psi
    \left (
        \frac{\mu_f(q) - \mu_f(q^+) - \epsilon}{\sigma_f(q)}
    \right ) 
    \\
    + \sigma_f(q)
    \psi
    \left (
        \frac{\mu_f(q) - \mu_f(q^+) -\epsilon}{\sigma_f(q)}
    \right )
    ,
\end{gathered}
\end{equation}
where $q^+$ is the maximizer of $\mu_f(q)$ over all $q \in \mathcal{Q}$, $\sigma_f(q) = \sqrt{\Sigma_f(q,q)}$ is the standard deviation, $\Psi$ is the standard normal cumulative distribution, $\psi$ is the standard normal density, and $\epsilon$ is an exploration hyperparameter.
The noisy version of EI is defined similarly:
\begin{equation}
\begin{gathered}
    \text{nEI}(q) = 
    (\mu_f(q) - \mu_f(q^+) - \epsilon)
    \Psi
    \left (
        \frac{\mu_f(q) - \mu_f(q^+) - \epsilon}
        {\hat{\sigma}_f (q)}
    \right )
    \\
    + \hat{\sigma}_f (q)
    \psi
    \left (
        \frac{\mu_f(q) - \mu_f(q^+) - \epsilon}{\hat{\sigma}_f (q)}
    \right )
    ,
\end{gathered}
\end{equation}
where $\hat{\sigma}_f (q) = \sqrt{\Sigma_f(q, q) + \Sigma_f(q^+, q^+) - 2*\Sigma_f(q, q^+)}$.
Finally, the Upper Confidence Bound is given by
\begin{equation}
    \text{UCB}(q) = \mu(q) + \epsilon \sigma_f(q)
    .
\end{equation}